\documentclass[lettersize,journal]{IEEEtran}
\usepackage{array}

\newcommand{\PreserveBackslash}[1]{\let\temp=\\#1\let\\=\temp}
\newcolumntype{C}[1]{>{\PreserveBackslash\centering}p{#1}}
\newcolumntype{R}[1]{>{\PreserveBackslash\raggedleft}p{#1}}
\newcolumntype{L}[1]{>{\PreserveBackslash\raggedright}p{#1}}
\usepackage{etoolbox}
\makeatletter
\patchcmd{\@makecaption}
  {\scshape}
  {}
  {}
  {}
\makeatother
\usepackage{amsmath,amsfonts}
\usepackage{algorithmic}
\usepackage{array}
\usepackage[colorlinks,
            linkcolor=black,       %%修改此处为你想要的颜色
            anchorcolor=black,  %%修改此处为你想要的颜色
            citecolor=black,   
            urlcolor = blue,
            ]{hyperref}
\usepackage{textcomp}
\usepackage{bbding} %首先在导言区调用bbding包
\usepackage{stfloats}
\usepackage{url}
\usepackage{verbatim}
\usepackage{graphicx}
\usepackage{subfigure}
\usepackage{color}
\usepackage{textcomp}
\usepackage{listings}

\def\BibTeX{{\rm B\kern-.05em{\sc i\kern-.025em b}\kern-.08em
    T\kern-.1667em\lower.7ex\hbox{E}\kern-.125emX}}
\makeatletter
\renewcommand{\@thesubfigure}{\hskip\subfiglabelskip}
\makeatother
\usepackage{balance}
\begin{document}
\title{Real HSI-MSI-PAN image dataset for the hyperspectral/multi-spectral/panchromatic image fusion and super-resolution fields}

\author{Shuangliang Li
        % <-this % stops a space
 \thanks{
 
 S. Li is with the Hubei Key Laboratory of Quantitative Remote Sensing of Land and Atmosphere, School of Remote Sensing and Information Engineering, Wuhan University, Wuhan, 430079, China and Hubei Luojia Laboratory, Wuhan University, Wuhan, 430079, China (e-mail: whu\_{lsl}@whu.edu.cn)

 }}

% \markboth{Journal of \LaTeX\ Class Files,~Vol.~18, No.~9, September~2020}
% {Unmixing based PAN guided fusion network for hyperspectral imagery}

\maketitle

% a novel training strategy on high resolution scale that benefit to the DDPM model's 迭代生成能力. 
% 一是融合结果模糊效应问题，即现有深度学习方法采用的损失函数和一步融合方式得到的结果会存在细节模糊问题，而DDPM模型可以通过迭代生成的方式得到具有真实细节的结果，可以有效缓解现有方法的模糊效应问题。这是为什么我们采用DDPM
% 二是融合结果空间和光谱保真度低的问题，即没有充分利用两幅输入影像中的空谱信息，我们提出了互补的空间和光谱DDPM双模型，更好地利用两幅输入影像信息，从输入两幅影像中充分提取空谱信息并进行融合。

\begin{abstract}
Nowadays, most of the hyperspectral image (HSI) fusion experiments are based on simulated datasets to compare different fusion methods. However, most of the spectral response functions and spatial downsampling functions used to create the simulated datasets are not entirely accurate, resulting in deviations in spatial and spectral features between the generated images for fusion and the real images for fusion. This reduces the credibility of the fusion algorithm, causing unfairness in the comparison between different algorithms and hindering the development of the field of hyperspectral image fusion. Therefore, we release a real HSI/MSI/PAN image dataset to promote the development of the field of hyperspectral image fusion. These three images are spatially registered, meaning fusion can be performed between HSI and MSI, HSI and PAN image, MSI and PAN image, as well as among HSI, MSI, and PAN image. This real dataset could be available at \href{https://aistudio.baidu.com/datasetdetail/281612}{https://aistudio.baidu.com/datasetdetail/281612}. The related code to process the data could be available at \href{https://github.com/rs-lsl/CSSNet}{https://github.com/rs-lsl/CSSNet} 
\end{abstract}


\section{Introduction}
% second, that GANs are able to trade off diversity
% for fidelity, producing high quality samples but not covering the whole distribution. We aim to bring
% these benefits to diffusion models, first by improving model architecture and then by devising a
% scheme for trading off diversity for fidelity.
% chikusei 解码器没有BN  资源解码器有BN 性能好

\IEEEPARstart {T}HE hyperspectral image (HSI) with hundreds of spectral bands provides rich spectral information for different materials and objects. And this has made it used in many real-world applications, such as classification \cite{ref1}\cite{ref2}, spectral unmixing \cite{ref7}\cite{refZheng0}, mineral exploitation \cite{ref6} and so on. 

However, due to the limitations of satellite imaging sensors, HSI always suffers from coarse spatial resolution, which leads to the loss of detailed information including tiny spatial textures and details. For example, the MODIS satellite captures the HSI at 500m resolution \cite{20292} and the Ziyuan (ZY) 1-02D satellite obtains the HSI at 30m resolution \cite{ref32}. To effectively improve the spatial resolution of the HSI, fusion methods between LRHSI and other auxiliary images have been explored greatly in recent years \cite{7050351}\cite{7000523}\cite{9076645}\cite{9625980}\cite{Bandara_2022_CVPR}.

% 事实上，还有其他一些方法可以提高高光谱影像的空间分辨率，如高光谱和多光谱影像融合，高光谱，多光谱和全色影像三者融合。上面这三种融合方法各有优劣，都是非常值得研究的方向，并且目前这些研究方向已经有相当数量的研究工作。
% 然而这些研究都是基于模拟数据集进行融合实验并对不同融合方法进行优劣比较。模拟数据集是指只能获取到一张原始高光谱影像，然后利用光谱响应函数和空间降采样函数对其进行光谱和空间降采样来生成模拟的高空间分辨率全色影像和低空间分辨率高光谱影像。然后对模拟生成的两幅影像进行融合以得到原始高分辨率高光谱影像。然而大部分研究所采用的光谱响应函数和空间降采样函数并不完全准确，使得生成的待融合影像和真实待融合影像存在空间和光谱特征的偏差。这降低了融合算法的可信度，使得不同算法之间的比较存在不公平性，阻碍了高光谱影像融合领域的发展。

Concretely, there are several fusion strategies, such as the fusion of HSI and panchromatic (PAN) image, the fusion of HSI and multi-spectral image (MSI), and the fusion of HSI, MSI, and PAN image. These three fusion methods each have their advantages and disadvantages, and they are all very worthy research directions. Currently, there is a considerable amount of research work in these directions, such as \cite{7050351}\cite{9676675}\cite{10108986}\cite{ref10}\cite{ref27}\cite{ref28}\cite{10041948}\cite{Quref}\cite{Bandara_2022_CVPR}\cite{9625980}\cite{9664535}\cite{9076645}\cite{refli}\cite{refdong}\cite{refguan1}\cite{10268922} in the fusion of HSI and PAN image field, \cite{10108986}\cite{9069930}\cite{li2023learning}\cite{10137388}\cite{5982386}\cite{Dian_2017_CVPR}\cite{8359412}\cite{8295275}\cite{8917657}\cite{8718504}\cite{9513426}\cite{9530284}\cite{8253497}\cite{8494792}\cite{Xie_2019_CVPR}\cite{9127776} in the fusion of HSI and MSI field, \cite{10557664}\cite{LI202230}\cite{9615043}\cite{jimaging4100118}\cite{10138912}\cite{6080924}
in the fusion of HSI, MSI and PAN image field.

%Recently, deep learning (DL)-based fusion methods have been developed greatly. This could be attributed to their superiority over traditional methods in terms of accuracy and efficiency, especially in the fields of image classification and fusion fields \cite{ref2}\cite{ref10}\cite{10108986}\cite{9069930}\cite{li2023learning}\cite{10137388}. 
However, these studies are mostly based on simulated datasets for fusion experiments and comparison of different fusion methods. A simulated dataset refers to obtaining only one original hyperspectral image, and then using spectral response functions and spatial downsampling functions to perform spectral and spatial downsampling on it to generate simulated high-spatial-resolution MSI/PAN image and low-spatial-resolution hyperspectral images. Then, these two simulated images are fused to obtain the high-spatial-resolution hyperspectral image. However, most of the adopted spectral response functions and spatial downsampling functions are not entirely accurate, resulting in deviations in spatial and spectral features between the generated images for fusion and the real images for fusion. This reduces the credibility of the fusion algorithm, causing unfairness in the comparison between different algorithms and hindering the development of the field of hyperspectral image fusion.

% 因此，本研究发布了一个高光谱真实数据集，包括高光谱/多光谱/全色影像。这三幅影像之间两两配准，即可以进行高光谱和多光谱融合，高光谱和全色融合，多光谱和全色融合，高光谱、多光谱和全色影像融合。在该真实数据家上进行融合实验可以消除选择的光谱/空间降采样函数存在偏差的问题（因为我们不需要这两个函数）。并且可以使得算法之间的比较更加公平，将极大地促进高光谱融合领域的发展。

Therefore, this study release a real HSI/MSI/PAN dataset, which includes hyperspectral, multi-spectral, and panchromatic images. These three images are spatially registered, meaning fusion can be performed between HSI and MSI, HSI and PAN image, MSI and PAN image, as well as among HSI, MSI, and PAN image. Conducting fusion experiments on this real dataset can eliminate the issue of bias in the selected spectral/spatial downsampling functions (since we do not need these two functions). Moreover, it allows for a fairer comparison between algorithms and will greatly promote the development of the field of hyperspectral image fusion.

% 接下来我们将列出该数据集的详细信息。包括空间分辨率和光谱分辨率以及影像大小。

\linespread{1.35}   % 表格行距
\begin{table}[t]
    \centering
    \setlength{\abovecaptionskip}{0.cm}% 和caption的间距
    \scriptsize   %调整表格内字体大小
    % \captionsetup{font={scriptsize}}
    \caption{Detailed information of the dataset}
    %设置列宽
    \setlength{\tabcolsep}{1.5mm}{
    \resizebox{\linewidth}{!}{
    \begin{tabular}{  C{2cm} C{2cm} C{2cm} C{2cm} C{1cm} C{1cm} }
\hline
image type&spatial resolution (m)&number of band& image size\\
\hline
HSI & 30 &	76 & [2071, 2033] \\
MSI & 10 &	8 & [6213, 6099] \\
PAN & 2.5 &	1 & [24852, 24396]\\
\hline
\end{tabular}}}
    \label{tab:dataset}
\end{table}

\begin{figure*}[t]
\centering
    \includegraphics[width=0.8\textwidth,trim=0 0 0 0,clip]{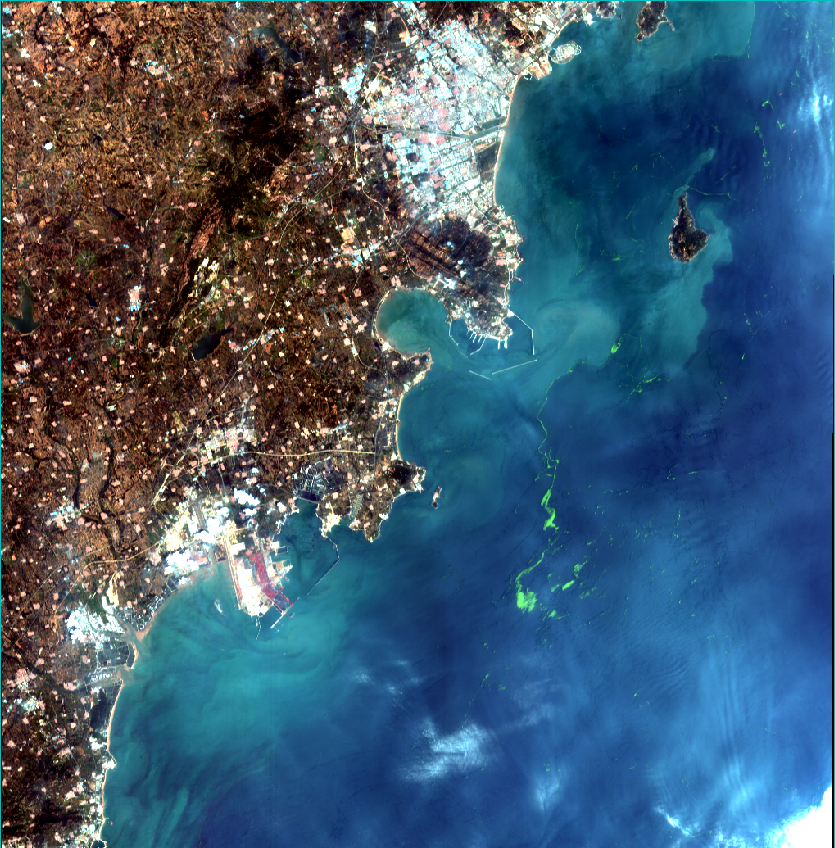}
    \caption{Hyperspectral image of ZY1-02D satellite. The RGB image is shown by utilizing the 35th, 16th, and 7th bands of the original HSI.}
    \label{fig:HSI_show}
\end{figure*}

\begin{figure*}[t]
\centering
    \includegraphics[width=0.8\textwidth,trim=0 0 0 0,clip]{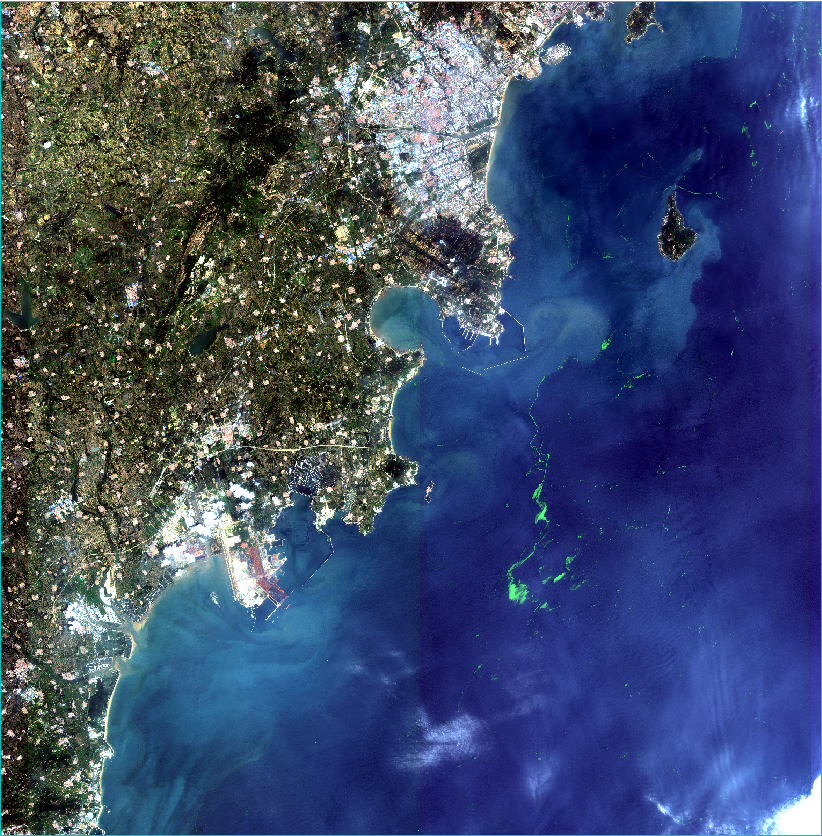}
    \caption{Multi-spectral image of ZY1-02D satellite. The RGB image is shown by utilizing the 3th, 2th, and 1th bands of the original MSI.}
    \label{fig:MSI_show}
\end{figure*}

\begin{figure*}[t]
\centering
    \includegraphics[width=0.8\textwidth,trim=0 0 0 0,clip]{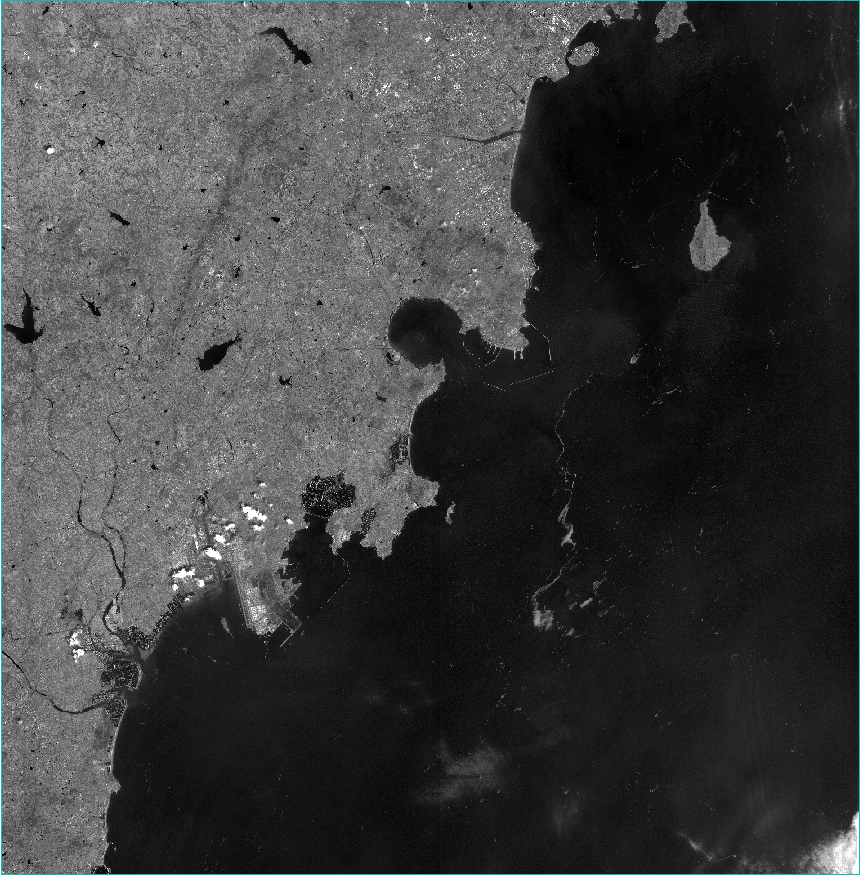}
    \caption{Panchromatic image of ZY1-02D satellite.}
    \label{fig:PAN_show}
\end{figure*}

This dataset was acquired by the Ziyuan-1 02D satellite and includes the HSI,MSI and PAN images. Their detailed information are listed in Table \ref{tab:dataset}. Note that the HSI originally has 166 bands and its spectral wavelength ranges from 395 to 2501 nm. Due to the low SNR in some long wavelength bands, we select a total of 76 spectral bands in visible and near-infrared wavelength range. 
For the fusion experiments, such as the fusion between HSI and PAN image, due to the lack of ground truth HRHSI, following the Wald protocol \cite{ref51}, we need to generate the spatially down-sampled LRHSI and PAN images as the input images of fusion model, and the original HSI is regarded as the reference image. During the testing phase, we can also input images at their original resolution into the network to obtain higher-resolution fused hyperspectral images. However, without a reference real image for comparison, we can only make a rough comparison of the spatial and spectral fidelity of the fused HSI with two input images. This spatial down-sampling fusion strategy is same as the fusion between HSI/MSI, MSI/PAN, HSI/MSI/PAN.

For the deep learning method, due to the large spatial size of this dataset, we need to crop this dataset into patches and input these patches into the fusion network. Note that in order to crop into patches, we need to remove the excess boundary pixels. Please see details in functions $generate\_data()$ and $crop\_data$ in $save\_image\_ziyuan\_hmp.py$ of \href{https://github.com/rs-lsl/CSSNet}{https://github.com/rs-lsl/CSSNet}.

{\small
\bibliographystyle{IEEEtran}
\bibliography{asme2e}

% Generated by IEEEtran.bst, version: 1.14 (2015/08/26)
\begin{thebibliography}{10}
\providecommand{\url}[1]{#1}
\csname url@samestyle\endcsname
\providecommand{\newblock}{\relax}
\providecommand{\bibinfo}[2]{#2}
\providecommand{\BIBentrySTDinterwordspacing}{\spaceskip=0pt\relax}
\providecommand{\BIBentryALTinterwordstretchfactor}{4}
\providecommand{\BIBentryALTinterwordspacing}{\spaceskip=\fontdimen2\font plus
\BIBentryALTinterwordstretchfactor\fontdimen3\font minus \fontdimen4\font\relax}
\providecommand{\BIBforeignlanguage}[2]{{%
\expandafter\ifx\csname l@#1\endcsname\relax
\typeout{** WARNING: IEEEtran.bst: No hyphenation pattern has been}%
\typeout{** loaded for the language `#1'. Using the pattern for}%
\typeout{** the default language instead.}%
\else
\language=\csname l@#1\endcsname
\fi
#2}}
\providecommand{\BIBdecl}{\relax}
\BIBdecl

\bibitem{ref2}
Q.~Zhu, W.~Deng, Z.~Zheng, Y.~Zhong, Q.~Guan, W.~Lin, L.~Zhang, and D.~Li, ``A spectral-spatial-dependent global learning framework for insufficient and imbalanced hyperspectral image classification,'' \emph{IEEE Transactions on Cybernetics}, pp. 1--15, 2021.

\bibitem{refZheng0}
K.~Zheng, L.~Gao, W.~Liao, D.~Hong, B.~Zhang, X.~Cui, and J.~Chanussot, ``Coupled convolutional neural network with adaptive response function learning for unsupervised hyperspectral super resolution,'' \emph{IEEE Transactions on Geoscience and Remote Sensing}, vol.~59, no.~3, pp. 2487--2502, 2021.

\bibitem{ref6}
N.~Yokoya, J.~C.-W. Chan, and K.~Segl, ``Potential of resolution-enhanced hyperspectral data for mineral mapping using simulated enmap and sentinel-2 images,'' \emph{Remote Sensing}, vol.~8, no.~3, 2016.

\bibitem{20292}
V.~Salomonson, W.~Barnes, P.~Maymon, H.~Montgomery, and H.~Ostrow, ``Modis: advanced facility instrument for studies of the earth as a system,'' \emph{IEEE Transactions on Geoscience and Remote Sensing}, vol.~27, no.~2, pp. 145--153, 1989.

\bibitem{ref32}
J.~Yu, D.~Liang, B.~Han, and H.~Gao, ``{Study on ground object classification based on the hyperspectral fusion images of ZY-1(02D) satellite},'' \emph{Journal of Applied Remote Sensing}, vol.~15, no.~4, p. 042603, 2021.

\bibitem{7050351}
Q.~Wei, N.~Dobigeon, and J.-Y. Tourneret, ``Bayesian fusion of multi-band images,'' \emph{IEEE Journal of Selected Topics in Signal Processing}, vol.~9, no.~6, pp. 1117--1127, 2015.

\bibitem{7000523}
M.~Simões, J.~Bioucas‐Dias, L.~B. Almeida, and J.~Chanussot, ``A convex formulation for hyperspectral image superresolution via subspace-based regularization,'' \emph{IEEE Transactions on Geoscience and Remote Sensing}, vol.~53, no.~6, pp. 3373--3388, 2015.

\bibitem{9076645}
Y.~Zheng, J.~Li, Y.~Li, J.~Guo, X.~Wu, and J.~Chanussot, ``Hyperspectral pansharpening using deep prior and dual attention residual network,'' \emph{IEEE Transactions on Geoscience and Remote Sensing}, vol.~58, no.~11, pp. 8059--8076, 2020.

\bibitem{9625980}
J.~Qu, S.~Hou, W.~Dong, S.~Xiao, Q.~Du, and Y.~Li, ``A dual-branch detail extraction network for hyperspectral pansharpening,'' \emph{IEEE Transactions on Geoscience and Remote Sensing}, vol.~60, pp. 1--13, 2022.

\bibitem{Bandara_2022_CVPR}
W.~G.~C. Bandara and V.~M. Patel, ``Hypertransformer: A textural and spectral feature fusion transformer for pansharpening,'' in \emph{Proceedings of the IEEE/CVF Conference on Computer Vision and Pattern Recognition (CVPR)}, June 2022, pp. 1767--1777.

\bibitem{9676675}
S.~Li, Y.~Tian, H.~Xia, and Q.~Liu, ``Unmixing-based pan-guided fusion network for hyperspectral imagery,'' \emph{IEEE Transactions on Geoscience and Remote Sensing}, vol.~60, pp. 1--17, 2022.

\bibitem{10108986}
S.~Li, Y.~Tian, C.~Wang, H.~Wu, and S.~Zheng, ``Hyperspectral image super-resolution network based on cross-scale nonlocal attention,'' \emph{IEEE Transactions on Geoscience and Remote Sensing}, vol.~61, pp. 1--15, 2023.

\bibitem{ref10}
L.~Loncan, L.~B. de~Almeida, J.~M. Bioucas-Dias, X.~Briottet, J.~Chanussot, N.~Dobigeon, S.~Fabre, W.~Liao, G.~A. Licciardi, M.~Simões, J.-Y. Tourneret, M.~A. Veganzones, G.~Vivone, Q.~Wei, and N.~Yokoya, ``Hyperspectral pansharpening: A review,'' \emph{IEEE Geoscience and Remote Sensing Magazine}, vol.~3, no.~3, pp. 27--46, 2015.

\bibitem{ref27}
L.~He, J.~Zhu, J.~Li, A.~Plaza, J.~Chanussot, and B.~Li, ``Hyperpnn: Hyperspectral pansharpening via spectrally predictive convolutional neural networks,'' \emph{IEEE Journal of Selected Topics in Applied Earth Observations and Remote Sensing}, vol.~12, no.~8, pp. 3092--3100, 2019.

\bibitem{ref28}
L.~He, J.~Zhu, J.~Li, D.~Meng, J.~Chanussot, and A.~Plaza, ``Spectral-fidelity convolutional neural networks for hyperspectral pansharpening,'' \emph{IEEE Journal of Selected Topics in Applied Earth Observations and Remote Sensing}, vol.~13, pp. 5898--5914, 2020.

\bibitem{10041948}
D.~He and Y.~Zhong, ``Deep hierarchical pyramid network with high- frequency -aware differential architecture for super-resolution mapping,'' \emph{IEEE Transactions on Geoscience and Remote Sensing}, vol.~61, pp. 1--15, 2023.

\bibitem{Quref}
J.~Qu, Z.~Xu, W.~Dong, S.~Xiao, Y.~Li, and Q.~Du, ``A spatio-spectral fusion method for hyperspectral images using residual hyper-dense network,'' \emph{IEEE Transactions on Neural Networks and Learning Systems}, pp. 1--15, 2022.

\bibitem{9664535}
W.~G.~C. Bandara, J.~M.~J. Valanarasu, and V.~M. Patel, ``Hyperspectral pansharpening based on improved deep image prior and residual reconstruction,'' \emph{IEEE Transactions on Geoscience and Remote Sensing}, vol.~60, pp. 1--16, 2022.

\bibitem{refli}
K.~Li, W.~Xie, Q.~Du, and Y.~Li, ``Ddlps: Detail-based deep laplacian pansharpening for hyperspectral imagery,'' \emph{IEEE Transactions on Geoscience and Remote Sensing}, vol.~57, no.~10, pp. 8011--8025, 2019.

\bibitem{refdong}
W.~Dong, Y.~Yang, J.~Qu, W.~Xie, and Y.~Li, ``Fusion of hyperspectral and panchromatic images using generative adversarial network and image segmentation,'' \emph{IEEE Transactions on Geoscience and Remote Sensing}, vol.~60, pp. 1--13, 2022.

\bibitem{refguan1}
P.~Guan and E.~Y. Lam, ``Multistage dual-attention guided fusion network for hyperspectral pansharpening,'' \emph{IEEE Transactions on Geoscience and Remote Sensing}, vol.~60, pp. 1--14, 2022.

\bibitem{10268922}
S.~Li, S.~Li, and L.~Zhang, ``Hyperspectral and panchromatic images fusion based on the dual conditional diffusion models,'' \emph{IEEE Transactions on Geoscience and Remote Sensing}, vol.~61, pp. 1--15, 2023.

\bibitem{9069930}
R.~Dian, S.~Li, and X.~Kang, ``Regularizing hyperspectral and multispectral image fusion by cnn denoiser,'' \emph{IEEE Transactions on Neural Networks and Learning Systems}, vol.~32, no.~3, pp. 1124--1135, 2021.

\bibitem{li2023learning}
S.~Li, R.~Dian, and H.~Liu, ``Learning the external and internal priors for multispectral and hyperspectral image fusion,'' \emph{Science China Information Sciences}, vol.~66, no.~4, p. 140303, 2023.

\bibitem{10137388}
R.~Dian, A.~Guo, and S.~Li, ``Zero-shot hyperspectral sharpening,'' \emph{IEEE Transactions on Pattern Analysis and Machine Intelligence}, vol.~45, no.~10, pp. 12\,650--12\,666, 2023.

\bibitem{5982386}
N.~Yokoya, T.~Yairi, and A.~Iwasaki, ``Coupled nonnegative matrix factorization unmixing for hyperspectral and multispectral data fusion,'' \emph{IEEE Transactions on Geoscience and Remote Sensing}, vol.~50, no.~2, pp. 528--537, 2012.

\bibitem{Dian_2017_CVPR}
R.~Dian, L.~Fang, and S.~Li, ``Hyperspectral image super-resolution via non-local sparse tensor factorization,'' in \emph{Proceedings of the IEEE Conference on Computer Vision and Pattern Recognition (CVPR)}, July 2017.

\bibitem{8359412}
S.~Li, R.~Dian, L.~Fang, and J.~M. Bioucas-Dias, ``Fusing hyperspectral and multispectral images via coupled sparse tensor factorization,'' \emph{IEEE Transactions on Image Processing}, vol.~27, no.~8, pp. 4118--4130, 2018.

\bibitem{8295275}
R.~Dian, S.~Li, A.~Guo, and L.~Fang, ``Deep hyperspectral image sharpening,'' \emph{IEEE Transactions on Neural Networks and Learning Systems}, vol.~29, no.~11, pp. 5345--5355, 2018.

\bibitem{8917657}
R.~Dian, S.~Li, L.~Fang, T.~Lu, and J.~M. Bioucas-Dias, ``Nonlocal sparse tensor factorization for semiblind hyperspectral and multispectral image fusion,'' \emph{IEEE Transactions on Cybernetics}, vol.~50, no.~10, pp. 4469--4480, 2020.

\bibitem{8718504}
R.~Dian and S.~Li, ``Hyperspectral image super-resolution via subspace-based low tensor multi-rank regularization,'' \emph{IEEE Transactions on Image Processing}, vol.~28, no.~10, pp. 5135--5146, 2019.

\bibitem{9513426}
J.~Xiao, J.~Li, Q.~Yuan, and L.~Zhang, ``A dual-unet with multistage details injection for hyperspectral image fusion,'' \emph{IEEE Transactions on Geoscience and Remote Sensing}, vol.~60, pp. 1--13, 2022.

\bibitem{9530284}
Y.~Zheng, J.~Li, Y.~Li, J.~Guo, X.~Wu, Y.~Shi, and J.~Chanussot, ``Edge-conditioned feature transform network for hyperspectral and multispectral image fusion,'' \emph{IEEE Transactions on Geoscience and Remote Sensing}, vol.~60, pp. 1--15, 2022.

\bibitem{8253497}
K.~Zhang, M.~Wang, S.~Yang, and L.~Jiao, ``Spatial–spectral-graph-regularized low-rank tensor decomposition for multispectral and hyperspectral image fusion,'' \emph{IEEE Journal of Selected Topics in Applied Earth Observations and Remote Sensing}, vol.~11, no.~4, pp. 1030--1040, 2018.

\bibitem{8494792}
C.~I. Kanatsoulis, X.~Fu, N.~D. Sidiropoulos, and W.-K. Ma, ``Hyperspectral super-resolution: A coupled tensor factorization approach,'' \emph{IEEE Transactions on Signal Processing}, vol.~66, no.~24, pp. 6503--6517, 2018.

\bibitem{Xie_2019_CVPR}
Q.~Xie, M.~Zhou, Q.~Zhao, D.~Meng, W.~Zuo, and Z.~Xu, ``Multispectral and hyperspectral image fusion by ms/hs fusion net,'' in \emph{Proceedings of the IEEE/CVF Conference on Computer Vision and Pattern Recognition (CVPR)}, June 2019.

\bibitem{9127776}
Z.~Wang, B.~Chen, R.~Lu, H.~Zhang, H.~Liu, and P.~K. Varshney, ``Fusionnet: An unsupervised convolutional variational network for hyperspectral and multispectral image fusion,'' \emph{IEEE Transactions on Image Processing}, vol.~29, pp. 7565--7577, 2020.

\bibitem{10557664}
Y.~Liu, X.~Cheng, H.~Li, S.~Luo, and B.~Yang, ``All in one: A unified network for hyperspectral image fusion,'' \emph{IEEE Transactions on Geoscience and Remote Sensing}, vol.~62, pp. 1--16, 2024.

\bibitem{LI202230}
\BIBentryALTinterwordspacing
K.~Li, W.~Zhang, D.~Yu, and X.~Tian, ``Hypernet: A deep network for hyperspectral, multispectral, and panchromatic image fusion,'' \emph{ISPRS Journal of Photogrammetry and Remote Sensing}, vol. 188, pp. 30--44, 2022. [Online]. Available: \url{https://www.sciencedirect.com/science/article/pii/S092427162200096X}
\BIBentrySTDinterwordspacing

\bibitem{9615043}
X.~Tian, W.~Zhang, Y.~Chen, Z.~Wang, and J.~Ma, ``Hyperfusion: A computational approach for hyperspectral, multispectral, and panchromatic image fusion,'' \emph{IEEE Transactions on Geoscience and Remote Sensing}, vol.~60, pp. 1--16, 2022.

\bibitem{jimaging4100118}
\BIBentryALTinterwordspacing
R.~Arablouei, ``Fusing multiple multiband images,'' \emph{Journal of Imaging}, vol.~4, no.~10, 2018. [Online]. Available: \url{https://www.mdpi.com/2313-433X/4/10/118}
\BIBentrySTDinterwordspacing

\bibitem{10138912}
X.~Tian, K.~Li, W.~Zhang, Z.~Wang, and J.~Ma, ``Interpretable model-driven deep network for hyperspectral, multispectral, and panchromatic image fusion,'' \emph{IEEE Transactions on Neural Networks and Learning Systems}, pp. 1--14, 2023.

\bibitem{6080924}
N.~Yokoya, T.~Yairi, and A.~Iwasaki, ``Hyperspectral, multispectral, and panchromatic data fusion based on coupled non-negative matrix factorization,'' in \emph{2011 3rd Workshop on Hyperspectral Image and Signal Processing: Evolution in Remote Sensing (WHISPERS)}, 2011, pp. 1--4.

\bibitem{ref51}
L.~Wald, T.~Ranchin, and M.~Mangolini, ``{Fusion of satellite images of different spatial resolutions: Assessing the quality of resulting images},'' \emph{{Photogrammetric engineering and remote sensing}}, vol.~63, no.~6, pp. 691--699, 1997.

\end{thebibliography}
}
% \end{comment}

\end{document}